\documentclass{bmvc2k}
\usepackage{subcaption}  

\usepackage{tikz}
\usepackage[inkscapelatex=false]{svg}
\usepackage{url}
\usepackage{amsmath,amssymb}
\usepackage{geometry}
\usepackage{xcolor}
\usepackage{colortbl}
\usepackage{pifont}
\usepackage{lipsum} 
\usepackage{graphicx} 
\usepackage{adjustbox} 
\usepackage{rotating}  
\usepackage{placeins}  
\usepackage{booktabs}
\usepackage{xspace}

\usepackage[switch]{lineno}

\usepackage{siunitx}
\usepackage{nicefrac}
\usepackage{xspace}
\usepackage{lineno}
\usepackage{nicefrac}
\usepackage{graphicx}
\usepackage{subcaption}
\usepackage[capitalize]{cleveref}
\crefname{section}{sec.}{secs.}
\Crefname{section}{Sec.}{Secs.}
\crefname{paragraph}{sec.}{secs.}
\Crefname{paragraph}{Sec.}{Secs.}
\crefname{table}{tab.}{tabs.}
\Crefname{table}{Tab.}{Tabs.}
\crefname{figure}{fig.}{figs.}
\Crefname{figure}{Fig.}{Figs.}
\crefname{equation}{eq.}{eqs.}
\Crefname{equation}{Eq.}{Eqs.}

\DeclareMathOperator*{\argmin}{arg\,min}

\makeatletter
\DeclareRobustCommand\onedot{\futurelet\@let@token\@onedot}
\def\@onedot{\ifx\@let@token.\else.\null\fi\xspace}
\newcommand{\etal}{et~al\onedot}
\newcommand{\eg}{e.g\onedot}

\title{End-to-End LiDAR optimization for 3D point cloud registration}

\addauthor{Siddhant Katyan}{siddhant.katyan.1@ulaval.ca}{1}
\addauthor{Marc-André Gardner}
{marc-andre.gardner@bentley.com}{2}
\addauthor{Jean-François Lalonde}
{jflalonde@gel.ulaval.ca}{1}

\addinstitution{
Université Laval \\
Quebec City, Canada
}
\addinstitution{
 Bentley Systems\\
 Quebec City, Canada
}

\runninghead{Siddhant Katyan et al.}{End-to-End LiDAR optimization for 3D PCR}

\def\eg{\emph{e.g}\bmvaOneDot}

\def\etal{\emph{et al}\bmvaOneDot}

\begin{document}

\maketitle

\begin{abstract}
LiDAR sensors are a key modality for 3D perception, yet they are typically designed independently of downstream tasks such as point cloud registration. Conventional registration operates on pre-acquired datasets with fixed LiDAR configurations, leading to suboptimal data collection and significant computational overhead for sampling, noise filtering, and parameter tuning. In this work, we propose an adaptive LiDAR sensing framework that dynamically adjusts sensor parameters, jointly optimizing LiDAR acquisition and registration hyperparameters. By integrating registration feedback into the sensing loop, our approach optimally balances point density, noise, and sparsity, improving registration accuracy and efficiency. Evaluations in the CARLA simulation demonstrate that our method outperforms fixed-parameter baselines while retaining generalization abilities, highlighting the potential of adaptive LiDAR for autonomous perception and robotic applications.
\end{abstract}


\section{Introduction}
\label{sec:introduction}

\begin{figure*}[!t]
\footnotesize
\centering
\includegraphics[width=1.0\linewidth]{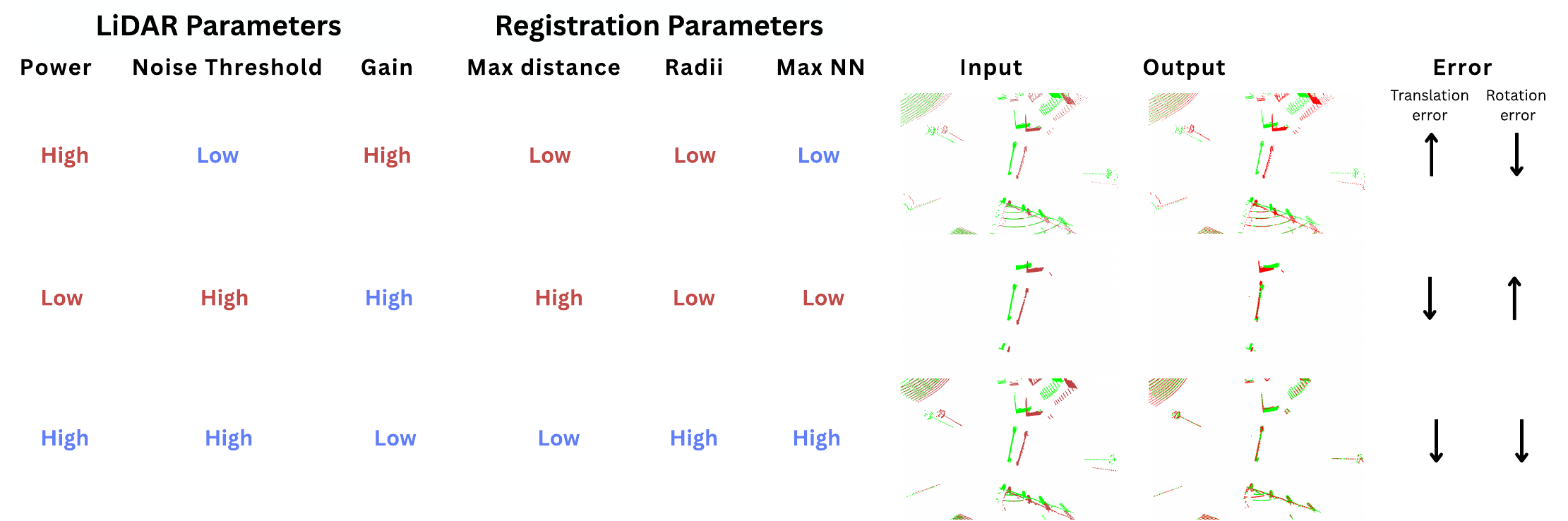}
\vspace{0.3em}
\caption{Variations in LiDAR scanning parameters and registration algorithm settings lead to divergent registration quality, demonstrating the sensitivity of the pipeline to manual tuning and the need for robust and adaptive end-to-end parameter optimization. Positive and negative interactions among parameters are highlighted in red and blue respectively.}
\label{fig:tuning}
\end{figure*}

Light Detection and Ranging (LiDAR) sensors have become an essential modality for 3D scene perception~\cite{lidar_object_detection} in applications such as autonomous navigation~\cite{lidar_autonomous_driving}, robotics~\cite{lidar_mobile_robotics}, and mapping~\cite{objectdetection_lidar, localization, mapping}. By emitting laser pulses and measuring their time-of-flight (ToF), LiDAR systems \cite{toflidarmobile}  generate dense 3D point clouds that accurately capture environmental geometry. Despite their widespread adoption, LiDAR sensors are typically designed as general-purpose data acquisition systems, with little to no adaptation to the specific perception tasks they support. We posit this may lead to suboptimal data collection strategies that can degrade the performance of downstream algorithms. One such task is point cloud registration, which aims to align multiple point clouds into a shared coordinate system. It serves as a key component in Simultaneous Localization and Mapping (SLAM)~\cite{slam_registration}, object tracking, and 3D reconstruction \cite{3Dreconstruction_registration}.

LiDAR sensing, much like camera-based imaging pipelines, involves a complex set of parameters that influence data quality and subsequent algorithmic performance \cite{prox_isp}. These parameters, spanning categorical, discrete, and continuous variables, include pulse power, gain, beam divergence, and detector sensitivity \cite{goudreault2023lidarhyperoptim}. Interactions between these factors are highly nonlinear, and suboptimal configurations can lead to increased noise, reduced effective resolution, and systematic distortions such as multiple reflections or occlusions. For instance, increasing laser power enhances signal-to-noise ratio (SNR) but can also introduce saturation effects, particularly on high-reflectivity surfaces. Similarly, excessive gain settings can amplify weak returns but at the cost of elevated background noise, which in turn affects geometric feature extraction critical for registration algorithms such as Point-to-Plane Iterative Closest Point (ICP)~\cite{pointplaneicp}.


In parallel, point cloud registration algorithms are highly sensitive to parameter selection, often requiring manual fine-tuning to adapt to different environments \cite{fgr}. Feature-based registration methods, such as Fast Point Feature Histograms (FPFH)~\cite{fpfh}, depend on well-calibrated search radii and voxel sizes, which are not universal across datasets \cite{registration_feature_tuning}.
Indeed, the effectiveness of global registration methods such as Fast Global Registration (FGR)~\cite{fgr} and MAC~\cite{mac} is contingent on selecting an optimal parameter set that aligns with both LiDAR sensor characteristics and scene complexity.

In \cref{fig:tuning}, we illustrate the motivation for jointly optimizing LiDAR sensing and registration algorithm (ICP) parameters by showing how manual tuning impacts registration accuracy across three cases. Case 1, with high \textit{power} and \textit{gain} but a low \textit{noise threshold}, produces a dense yet noisy point cloud; combined with small \textit{max distance}, \textit{radii}, and \textit{max nn}, this results in poor normal estimation, weak correspondence rejection, and large errors. Case 2, using low \textit{power} and a high \textit{noise threshold}, generates a sparse but clean point cloud; however, small \textit{radii}, \textit{max nn}, and large \textit{max distance} permit spurious correspondences, leading to noticeable registration drift. Case 3, with high \textit{power} and a high \textit{noise threshold}, yields a clean, moderately dense point cloud; paired with larger \textit{radii}, \textit{max nn}, and conservative \textit{max distance}, it achieves the most accurate alignment. Together, these cases highlight that robust registration requires context-aware, scene-specific parameter tuning, while poor configurations lead to degraded alignment.

In this paper, we propose a joint optimization framework that integrates LiDAR acquisition parameter tuning with registration algorithm hyperparameter selection. Instead of treating LiDAR sensing and registration as independent processes as is typically done, our approach introduces an adaptive feedback mechanism wherein registration performance guides LiDAR parameter adjustments. This enables task-aware sensing, wherein LiDAR settings such as pulse power, scanning pattern, and gain are adapted based on scene characteristics and registration performance metrics.

We implement and evaluate our method using the CARLA simulation engine \cite{carla} and the LiDAR simulation framework of Goudreault~\etal~\cite{goudreault2023lidarhyperoptim}. 
We perform experiments on synthetically-generated datasets mimicking real captures (with added gaussian noise to replicate real world sensor noise), in three different kinds of environments: structured, semi-structured, and unstructured. Our results demonstrate that joint optimization of LiDAR and registration settings significantly enhances registration accuracy and robustness across diverse environments compared to default configurations, bringing up to $3\times$ improvement in important metrics such as registration recall. We also demonstrate the generalization ability of our approach by showing how a set of parameters optimized on just a few scenes can generalize and improve performance on any \textit{similar} scene (e.g., unstructured), even those taken at a different location. These findings underscore the importance of adaptive LiDAR tuning in enhancing point cloud registration for real-world applications.

%
%
%
%

\section{Related work}
\label{sec:relwork}

\subsection{LiDAR and registration hyperparameter optimization}

LiDAR DSP involves multiple stages, including signal denoising, waveform processing, and point cloud generation. Conventional approaches treat the LiDAR sensor as a static system with fixed acquisition parameters. Several studies have explored the impact of tuning sensor parameters such as pulse power, receiver gain, and detection threshold on LiDAR performance. For instance, Goudreault et al. \cite{goudreault2023lidarhyperoptim} introduced a hyperparameter optimization framework for LiDAR DSP to improve sensing quality in adverse conditions. Their approach demonstrated that adaptive parameter selection enhances point cloud quality, particularly in scenes with varying reflectivity. However, their work did not explicitly consider the downstream task of point cloud registration, leaving an open question on how sensor-level optimization impacts registration accuracy.

In another study, Selmer~\etal~\cite{physics_lidar} proposed a physics-based LiDAR simulation framework that models sensor noise and material reflectance properties. While their model provides insights into LiDAR behavior under different settings, it remains computationally expensive for real-time applications. Other works \cite{denoising_lidar} have explored machine learning-based LiDAR denoising techniques, but these methods primarily focus on reducing noise rather than optimizing the LiDAR parameters themselves.

A key limitation in current LiDAR DSP research is the lack of task-aware parameter tuning. Existing methods optimize LiDAR parameters for generic data quality metrics, but do not integrate feedback from downstream perception tasks, such as point cloud registration or SLAM. This motivates the need for a closed-loop LiDAR parameter optimization framework that dynamically adjusts sensor settings based on real-time registration performance.

\subsection{Point cloud registration}

Point cloud registration is a fundamental problem in 3D vision, with applications in mapping, localization, and scene reconstruction. Classical registration algorithms such as Iterative Closest Point (ICP) \cite{pointicp} and Fast Global Registration (FGR) \cite{fgr} require careful tuning of hyperparameters like voxel size, normal estimation radius, and correspondence rejection thresholds. These parameters heavily influence registration accuracy, yet they are often selected heuristically or manually tuned for specific datasets.

Recent works have attempted to automate hyperparameter selection for point cloud registration. Guo~\etal~\cite{adaptive_icp} proposed an adaptive feature weighting scheme for improving registration robustness, while Zhou~\etal~\cite{correspondence_filtering_registration} studied deep learning-based correspondence filtering to reduce registration errors. However, these methods still assume static LiDAR data acquisition, neglecting the role of LiDAR sensing parameters in shaping point cloud characteristics.

Another line of research explores scene-dependent registration parameter tuning. For example, Ao~\etal~\cite{buffer} analyzed how point cloud registration methods perform under varying point cloud densities and noise levels, balancing accuracy,
efficiency, and generalizability. This aligns with our motivation to develop an adaptive LiDAR sensing and registration framework, where sensor parameters are co-optimized alongside registration hyperparameters.

\begin{figure*}[!t]
\footnotesize
\centering
\begin{tabular}{c}
\includegraphics[width=1.0\linewidth]{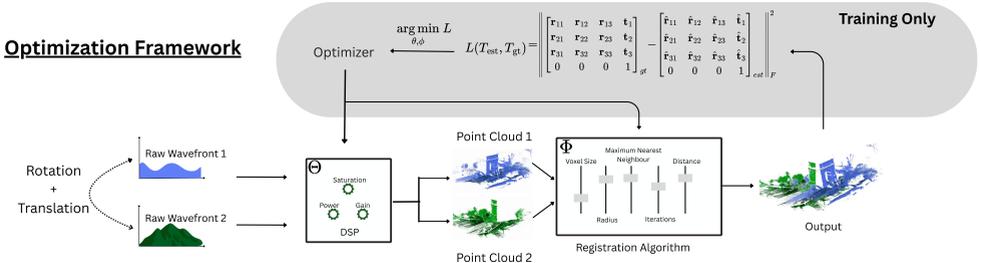}
\end{tabular}
\vspace{.5em}
\caption{LiDAR optimization framework for point cloud registration optimization. $\mathbf{T}_\text{gt}$ and $\mathbf{T}_\text{est}$ are ground-truth matrix and estimated transform respectively for point cloud registration.}
\label{fig:lidar_pc_optimization}
\vspace{1em}
\end{figure*}

\section{Optimization framework}
\label{sec:optimization}

Our optimization framework jointly optimizes LiDAR and registration parameters through a unified covariance matrix adaptation evolution strategy (CMA-ES) \cite{cmaes}. 

\subsection{Optimization objective}

Formally, we wish to obtain the best set of LiDAR $\Theta_\text{lidar}$ and registration $\Theta_\text{reg}$ parameters such that performance at the registration task is maximized for a given dataset. We first define the ``LiDAR'' function 
\begin{equation}
\mathbf{O} = f_\text{lidar}(\mathbf{P}, \mathcal{S}; \Theta_\text{lidar}) \,,
\label{eq:lidar}
\end{equation}
which accepts as input a 6-DoF pose $\mathbf{P}$ (represented as a $4 \times 4$ matrix) of the LiDAR in a given scene $\mathcal{S}$, and produces as output the corresponding point cloud $\mathbf{O}$. Here, $f_\text{lidar}$ is parametrized by $\Theta_\text{lidar}$. 

We also define the ``registration'' function
\begin{equation}
\mathbf{T} = f_\text{reg}(\mathbf{O}_1, \mathbf{O}_2; \Theta_\text{reg}) \,,
\label{eq:reg}
\end{equation}
which accepts two points clouds $\mathbf{O}_1$ and $\mathbf{O}_2$ as input, and returns the pose transform $\mathbf{T}$ s.t. $\mathbf{T} \mathbf{O}_2$ is aligned with $\mathbf{O}_1$. Similarly to \cref{eq:lidar}, $f_\text{reg}$ is parametrized by $\Theta_\text{reg}$. 

Our goal is to find the best set of $(\Theta_\text{lidar}, \Theta_\text{reg})$ such that successful registration can be attained for a set of pose pairs $\mathcal{P} = \{(\mathbf{P}_1, \mathbf{P}_2), (\mathbf{P}_2, \mathbf{P}_3), ...\}$ in a given scene $\mathcal{S}$. In practice, pairs of locations are set a certain distance apart from each other, \eg, 5 meters. Assuming knowledge of a training set, containing a set of pairs of camera poses and their corresponding transformation, we define the following optimization objective:
\begin{equation}
\argmin_{\Theta_\text{lidar}, \Theta_\text{reg}} \sum_{\mathbf{P}_i,\mathbf{P}_j} ||f_\text{opt}(\mathbf{P}_i, \mathbf{P}_j; \Theta_\text{lidar}, \Theta_\text{reg}) - \mathbf{T}^*_{i,j}||_F \,,
\label{eq:opt}
\end{equation}
where the ground truth transformation between the two point clouds are obtained by their relative pose $\mathbf{T}^*_{i,j} = \mathbf{P}_i\mathbf{P}_j^{-1}$. $f_\text{opt}$ is defined as
\begin{equation}
f_\text{opt} \equiv f_\text{reg}(f_\text{lidar}(\mathbf{P}_i; \Theta_\text{lidar}), f_\text{lidar}(\mathbf{P}_j; \Theta_\text{lidar}); \Theta_\text{reg}) \,,
\label{eq:f-opt}
\end{equation}
which employs the LiDAR function $f_\text{lidar}$ to obtain the corresponding point clouds at poses $\mathbf{P}_i$ and $\mathbf{P}_j$, and attempts to register them with the registration function $f_\text{reg}$. Both of these functions are defined in greater details next. 

\subsection{Defining the LiDAR function}

We rely on the LiDAR simulation model from \cite{goudreault2023lidarhyperoptim} to define the LiDAR function $f_\text{lidar}$. This generates full transient waveforms by extracting scene response, ambient light and object reflectances generated by the CARLA driving simulator~\cite{carla}. The waveforms are then processed by the LiDAR DSP, resulting in a point cloud $\mathbf{O}$. 

The LiDAR DSP parameters $\Theta_\text{lidar}$ govern the analog-to-digital conversion and peak detection pipeline that transforms raw analog wavefronts $\psi(t)$ into discrete 3D points. The power parameter $P$ controls the analog gain stage, directly influencing the signal-to-noise ratio (SNR) of detected returns. This interacts nonlinearly with the noise floor threshold $N$, which determines the minimum detectable signal amplitude through the conditional
\begin{equation}
\text{PeakDetected} = \begin{cases}
1 & \text{if } \max(\psi(t)) > N \\
0 & \text{otherwise}
\end{cases} \,.
\end{equation}
The waveform saturation level $S$ digitally clips the maximum return intensity to prevent sensor blooming, while the pulse width $W$ affects both the minimum resolvable distance and the range resolution
\begin{equation}
\Delta R = \frac{c \cdot W}{2} \,,
\end{equation}
where $c$ is the speed of light. These parameters collectively determine the point cloud density, noise characteristics, and effective range through the processing pipeline. 

In our experiments, we used the parameters and recommended ranges suggested by \cite{goudreault2023lidarhyperoptim}.

\subsection{Defining the registration function}

The registration pipeline is governed by parameters $\Theta_\text{reg}$, which encapsulate both the parameters of the registration algorithm and those of the feature descriptor (FPFH~\cite{fpfh} in this case). The process begins with voxel grid downsampling at resolution $\phi_v \in [0.01, 0.1]\text{m}$, which balances geometric detail preservation and computational efficiency. Feature extraction then proceeds using FPFH, parameterized by normal estimation radius $\phi_{\text{rad1}} \in [0.01, 0.1]\text{m}$, feature computation radius $\phi_{\text{rad2}} \in [\phi_{\text{rad1}}, 1.0]\text{m}$, and nearest neighbor count $\phi_{\text{knn}} \in \mathbb{Z}^+$ (typically 10--100). The radius $\phi_{\text{rad1}}$ determines the local region for computing surface normals, with smaller values capturing fine detail and larger values improving robustness. $\phi_{\text{rad2}}$ defines the neighborhood for aggregating Simplified Point Feature Histograms (SPFHs), enabling broader contextual information in FPFH descriptors. The constraint $\phi_{\text{rad2}} \geq \phi_{\text{rad1}}$ ensures that normals are computed at a finer scale than feature aggregation. The number of neighbors $\phi_{\text{knn}}$ trades off local detail sensitivity and descriptor stability, impacting both accuracy and the $O(N\phi_{\text{knn}})$ complexity of feature computation. Note that tuning feature extractor parameters is essential for both classical and deep learning–based registration approaches, as it directly affects the accuracy of the registration algorithm.

Two critical tuning parameters for the registration stage are the maximum correspondence distance $\phi_d \in [0.1, 2.0]$ m and the iteration count $\phi_{\text{icp}}$. $\phi_d$ sets a threshold for valid correspondences.
%
%
If $\phi_d$ is too small, insufficient correspondences may degrade convergence; if too large, outliers may dominate. The iteration count $\phi_{\text{fgr}}$ governs convergence behavior in FGR \cite{fgr}.  Together, $\Theta_{\text{reg}} = \{\phi_v, \phi_{\text{rad1}}, \phi_{\text{rad2}}, \phi_{\text{knn}}, \phi_d, \phi_{\text{fgr}}\}$ jointly govern feature extraction and registration. 

\subsection{Optimization procedure}

The CMA-ES optimizer is employed to solve the optimization problem from \cref{eq:opt}, which depends on both the LiDAR and registration parameters $\Theta_\text{lidar}$ and $\Theta_\text{reg}$ respectively. The optimizer handles these complex interactions of parameters through its adaptive covariance matrix update, where the off-diagonal elements in $\mathbf{C}_t$ automatically capture parameter correlations discovered during optimization. This allows the algorithm to identify, for instance, that increasing LiDAR power might require adjusting the FPFH radii to maintain optimal feature distinctiveness.

We adhere to the physical constraints of the hardware and set the bounds of the tuning parameters of LiDAR and registration in accordance with the range mentioned in \cite{goudreault2023lidarhyperoptim, open3dmodernlibrary3d}. The parameters are normalized between $(0,1)$ for better stability of the CMAES algorithm and are unnormalized during the inference through the DSP and registration. The convergence criteria is set to max number of generations or threshold translation error and rotation error
\begin{equation}
e_\mathbf{t} = \|\mathbf{t}_\text{gt} - \mathbf{t}_\text{est}\|_2 \; \text{and} \;
e_\mathbf{R} = \cos^{-1}\left(\frac{\text{tr}(\mathbf{R}_\text{gt}^\top \, \mathbf{R}_\text{est}) - 1}{2}\right) \,,
\end{equation}
where $\text{tr}(\cdot)$ is the trace. 

\section{Experimental validation}
\label{sec:experiments}

\subsection{Data collection}

We evaluate our proposed approach on data generated by the LiDAR simulation framework of \cite{goudreault2023lidarhyperoptim}, which itself leverages the CARLA engine~\cite{carla}, which we modified to enable free waypoint-based positioning. This allows us to simulate diverse LiDAR configurations beyond those constrained by vehicle-mounted placements. The data was captured in various towns in CARLA simulation environment featuring both urban and rural settings with buildings, roads, trees, and other vegetation. The virtual LiDAR sensor was positioned at a height of 2 meters above the ground with a horizontal and vertical view range of $+90^\circ$ to $-90^\circ$.

Similar to the ETH Zurich dataset~\cite{eth_dataset} standard, we categorize the environment into structured, semi-structured, and unstructured scenes. This classification allows us to assess the effectiveness of scene-dependent LiDAR tuning under varying environmental complexities. Locations in CARLA scenes corresponding to these three types of environments were identified manually. We render $10$ pairs of scans for each of the environment types, collected in three different CARLA towns. A test set of $100$ pairs is also generated in the same manner, but in different locations such that there is no overlap between train and test viewpoints.

For both train and test set, we follow the benchmarking methodology in \cite{benchmark_pcr} and generate raw waveforms pairs, in which the second waveform is acquired at a distance of 5 meters from the first one. Dataset generation for local registration methods follows the same protocol, but with a distance of 2 meters instead of 5.

\subsection{Training details}

We train and evaluate our approach on raw waveforms pairs. Training is performed on each category of scenes (structured, semi-structured, unstructured) separately. 

Our approach can optimize the LiDAR parameters, the registration parameters, or both. At training time, we initialize the LiDAR parameters using the default values from \cite{goudreault2023lidarhyperoptim} and the registration parameters from Open3D defaults \cite{open3dmodernlibrary3d}. We then run a CMA-ES optimization using as many individuals as the number of parameters to optimize, which we let run for 30 generations. In line with CMA-ES best practices \cite{cmaes}, we normalize the parameter values in the $[0, 1]$ range. Training time varies depending on the number of parameters optimized, but typically takes between 24 and 30 hours on a Ryzen 3700X (8 cores).

During training, CMA-ES individuals receive as fitness the Frobenius norm of the difference between the ground truth transformation matrix and the one predicted by the registration technique, averaged over the 10 scenes (see \ref{eq:opt}). No outlier removal is applied prior to this average.

\subsection{Evaluation methodology}

We evaluate our approach using both local and global registration algorithms. Specifically, we evaluate \textit{Point-to-Plane ICP (ICP)} as an example of a local algorithm. For global algorithms, we experiment with \textit{Fast Global Registration (FGR)}~\cite{fgr}, an optimization-based method, and \textit{3D Registration with Maximal Cliques (MAC)}~\cite{mac}, a state-of-the-art clique-based graph matching algorithm.

At evaluation time, we use the best parameters set, defined as the one obtaining the best registration performance on the train set, and apply those parameters to generate registered point cloud pairs from the raw waveforms. Following \cite{lim2023quatro-plusplus, gpicp, photogrammetric_registration_dataset}, we report registration recall as the percentage of pairs for which the registration is considered successful. For the global registration methods, we use thresholds of 1 meter and $5^\circ$ for the translation and rotation errors, respectively. For ICP, these thresholds are lowered to $0.5$ meter and $2.5^\circ$. We also report the average translation and rotation errors on the successful registrations.




\begin{table*}[t]
    \centering
    \footnotesize
    \setlength{\tabcolsep}{3pt}
    \newcommand{\optyes}{\textcolor{green}{\ding{51}}\xspace}
    \newcommand{\optno}{\textcolor{red}{\ding{55}}\xspace}

    \caption{Performance comparison of tuning LiDAR parameters in conjunction with registration algorithm parameters across various environments on the test data. A check (\optyes) indicates that the corresponding component was tuned, while a cross (\optno) means it was kept fixed. Best results are highlighted in green.}
    \vspace{0.6em}
    \label{tab:results1}

    \begin{tabular}{llllccccccccc}
        \toprule
        & & & & 
        \multicolumn{3}{c}{\textbf{Structured env.}} & 
        \multicolumn{3}{c}{\textbf{Semi-structured env.}} & 
        \multicolumn{3}{c}{\textbf{Unstructured env.}} \\
        & & & & Trans. & Rot. & Recall & Trans. & Rot. & Recall & Trans. & Rot. & Recall \\
        \midrule
        \multicolumn{4}{l}{\textbf{Local Registration}} \\
        \optno & LiDAR & \optno & ICP & 3cm & 0.2\textdegree & 100 & 4cm & 0.3\textdegree& 100 & 4cm & 0.4\textdegree  & 100 \\
        \optyes & LiDAR & \optno & ICP & 3cm & 0.2\textdegree & 100 & 4cm & 0.3\textdegree& 100 & 4cm & 0.4\textdegree  & 100 \\
        \optno & LiDAR & \optyes & ICP & 2cm & 0.2\textdegree & 100 & 6cm & 0.6\textdegree & 100 & 10cm & 0.9\textdegree & 100 \\
        \optyes & LiDAR & \optyes & ICP & \cellcolor{green!20} 1cm & \cellcolor{green!20} 0.1\textdegree & 100 & \cellcolor{green!20} 2cm & \cellcolor{green!20} 0.2\textdegree & 100 & \cellcolor{green!20} 2cm & \cellcolor{green!20} 0.1\textdegree & 100 \\
        \midrule
        \multicolumn{4}{l}{\textbf{Global Registration}} \\
        \optno & LiDAR & \optno & FGR & 61cm & 3.33\textdegree & 14.1 & 69cm & 3.41\textdegree & 6.06 & 67cm & 2.64\textdegree & 18.1 \\
        \optyes & LiDAR & \optno & FGR & 78cm & 2.75\textdegree & 22.2 & 69cm & 3.22\textdegree & 11.1 & \cellcolor{green!20}17cm & 2.59\textdegree & 5.05 \\
        \optno & LiDAR  & \optyes & FGR & 61cm & 2.90\textdegree & 31.3 & 80cm & \cellcolor{green!20} 2.43\textdegree & 13.1 & 63cm & 3.12\textdegree & 35.3 \\
        \optyes & LiDAR & \optyes & FGR & 63cm & \cellcolor{green!20} 1.73\textdegree &\cellcolor{green!20} 35.4 & 78cm & 2.8\textdegree &\cellcolor{green!20} 17.1 & 71cm & \cellcolor{green!20} 1.71\textdegree & \cellcolor{green!20}60.0 \\
        & & & & & & & & & & & & \\*[-0.01em]
        \optno & LiDAR & \optno & MAC & 87cm & 4.02\textdegree & 48.4 & 92cm & 4.61\textdegree & 49.4 & 61cm & 3.61\textdegree & 56.6 \\
        \optyes & LiDAR & \optno & MAC & 94cm & 4.46\textdegree & \cellcolor{green!20} 56.6 & 62cm & 3.12\textdegree & 51.5 & 51cm & 2.92\textdegree & 60.0 \\
        \optno & LiDAR & \optyes & MAC & 65cm & 2.81\textdegree & 48.4 & 59cm & 2.98\textdegree & 47.5 & 49cm & 2.98\textdegree & 60.0 \\
        \optyes & LiDAR & \optyes & MAC & \cellcolor{green!20}56cm & \cellcolor{green!20}2.63\textdegree &  52.6 & \cellcolor{green!20}51cm & \cellcolor{green!20}2.46\textdegree & \cellcolor{green!20} 68.6 & \cellcolor{green!20}41cm &  \cellcolor{green!20}2.63\textdegree & \cellcolor{green!20} 81.8 \\
        \bottomrule
    \end{tabular}
\end{table*}

    
    
    

    

\begin{figure*}[!t]
\footnotesize
\centering
\includegraphics[width=1.0\linewidth]{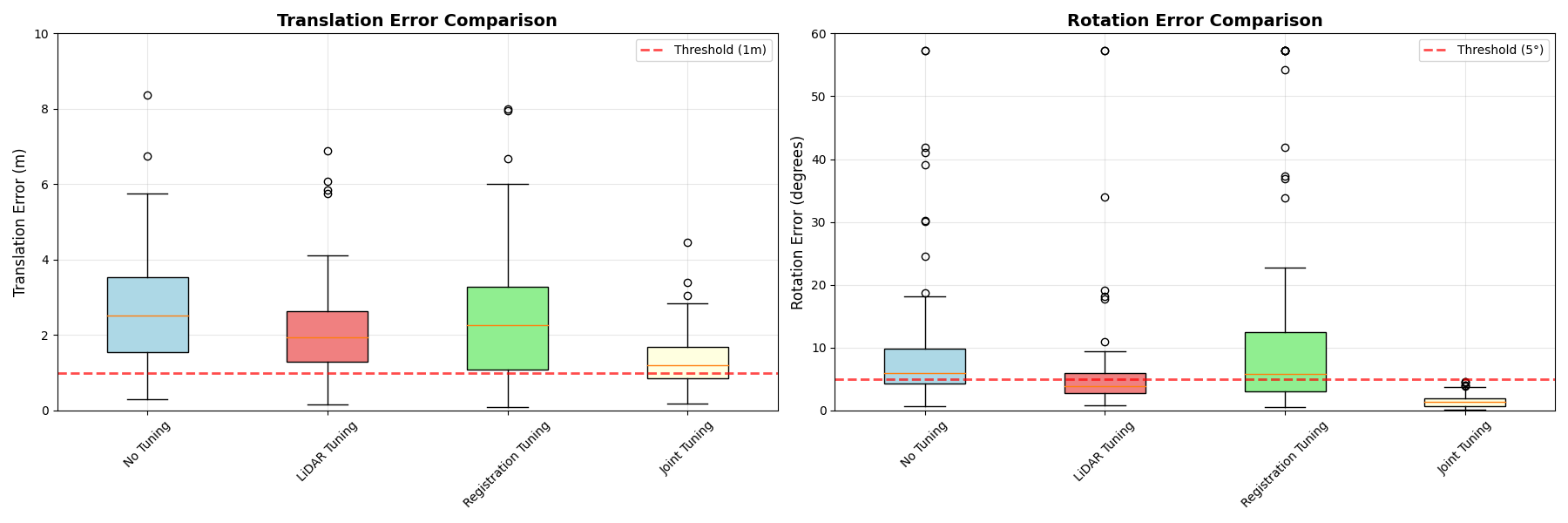}
\vspace{0.1em}
\caption{Error distribution on the test data including failure registration cases for FGR structured environment.}
\label{fig:robustness_plot}
\end{figure*}

\subsection{Experimental results}

\Cref{tab:results1} presents a comparative evaluation of different tuning protocols involving LiDAR and registration algorithm parameters on the test set. We consider four settings, indicated graphically in the figure: the ``default configuration'', where standard default values (\eg, taken from \cite{goudreault2023lidarhyperoptim} for LiDAR) are used for both LiDAR and registration; ``LiDAR-only tuning'', where LiDAR parameters are optimized while keeping the registration algorithm fixed; ``registration-only tuning'', where the registration algorithm is tuned while keeping the default LiDAR setup; and finally, ``joint tuning'', where LiDAR and registration parameters are jointly optimized in an end-to-end fashion to enable scene-adaptive performance. These evaluations are carried out for both local and global registration algorithms.

ICP (local) obtains perfect recall across the entire test set, whether we optimize its parameters or not. However, our end-to-end optimization results in an improvement for the average translation and rotation errors, particularly when we jointly optimize LiDAR and ICP parameters. Using the parameters coming from this joint optimization results in the lowest errors for all environment types. For FGR and MAC (global), our joint LiDAR+method optimization significantly improves the registration recall in all scenarios when compared to the use of baseline parameters, except for MAC applied on structured scenes where it comes second. For instance, with unstructured environments, FGR recall jumps from 18\% to 60\%. Moreover, inliers translation and rotation errors are also globally improved.

These results also highlight the benefit of parameter optimization for registration methods: optimizing solely the FGR parameters, without touching LiDAR acquisition parameters, already outperforms the baseline by a $2\times$ improvement factor on registration recall. In other words, our approach may improve results even for the more common scenarios where point clouds are already captured and there is no LiDAR ``in-the-loop''.

\Cref{fig:robustness_plot} highlights the robustness of joint tuning: the distribution of the translation and rotation errors over the entire 100 test set pairs show that jointly optimizing LiDAR and registration parameters yields smaller error ranges and fewer failures compared to other tuning strategies, even on scenes that were not part of the training, merely \textit{similar in content}.

Overall, these results, illustrated qualitatively in \cref{fig:qualitative_results}, demonstrate our approach generalization ability and confirm our hypothesis: optimizing LiDAR and registration parameters on a few scenes yields a parameters set which generalizes to \textit{every similar scene}, not only the ones captured at the same or a close location. As a result, the time required for training/optimization is of less importance, since a single optimization run results in a set of parameters which may be used on other, new similar scenes.

\newcommand{\imshowfgr}[1]{
\frame{\includegraphics[width=0.22\linewidth]{#1}}\llap{\raisebox{0.8cm}{\frame{\includegraphics[width=0.1\linewidth, trim=9cm 5cm 9cm 5cm, clip]{#1}}}\hspace{1.7cm}}}
\newcommand{\imshowmac}[1]{
\frame{\includegraphics[width=0.22\linewidth]{#1}}\llap{\raisebox{0.81cm}{\frame{\includegraphics[width=0.1\linewidth, trim=8.25cm 5cm 8.25cm 5cm, clip]{#1}}}\hspace{1.7cm}}}
\begin{figure*}
\footnotesize
\setlength{\tabcolsep}{3pt}
\centering
    \begin{tabular}{ccccc}
    & Input point clouds & LiDAR-only opt. & Registration-only opt. & Joint opt. \\
    \rotatebox{90}{\hspace{2.2em}FGR} & 
    \imshowfgr{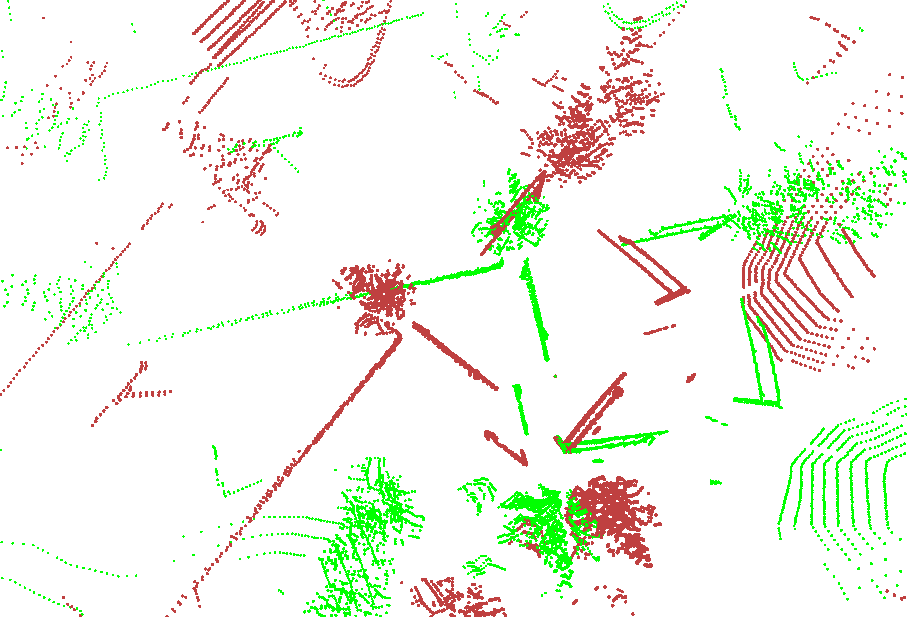} & 
    \imshowfgr{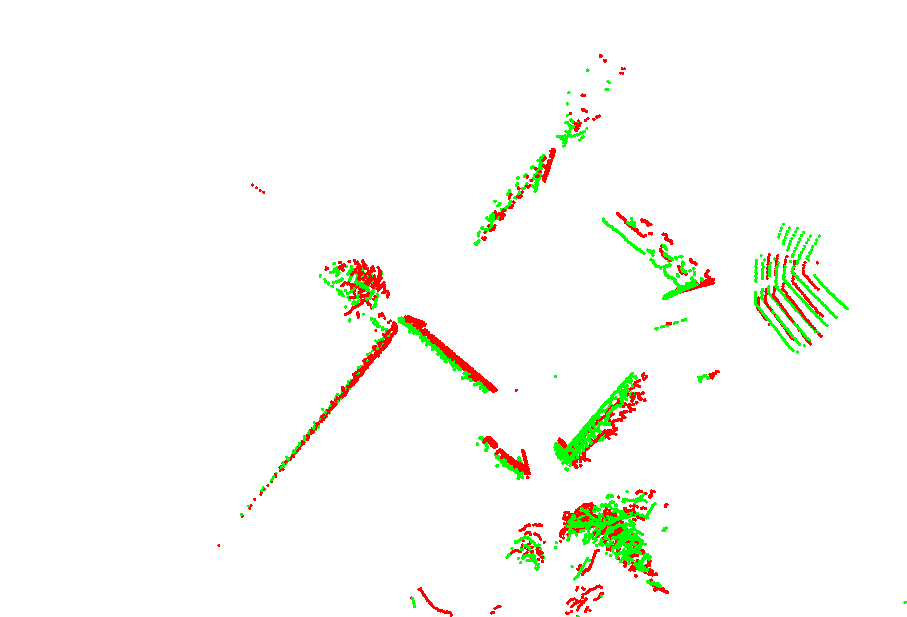} &
    \imshowfgr{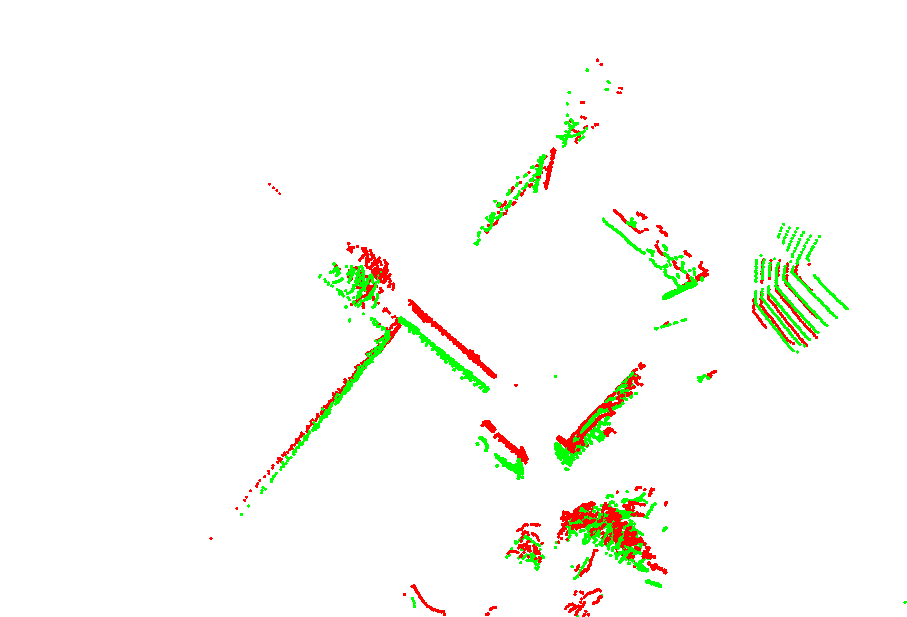} & 
    \imshowfgr{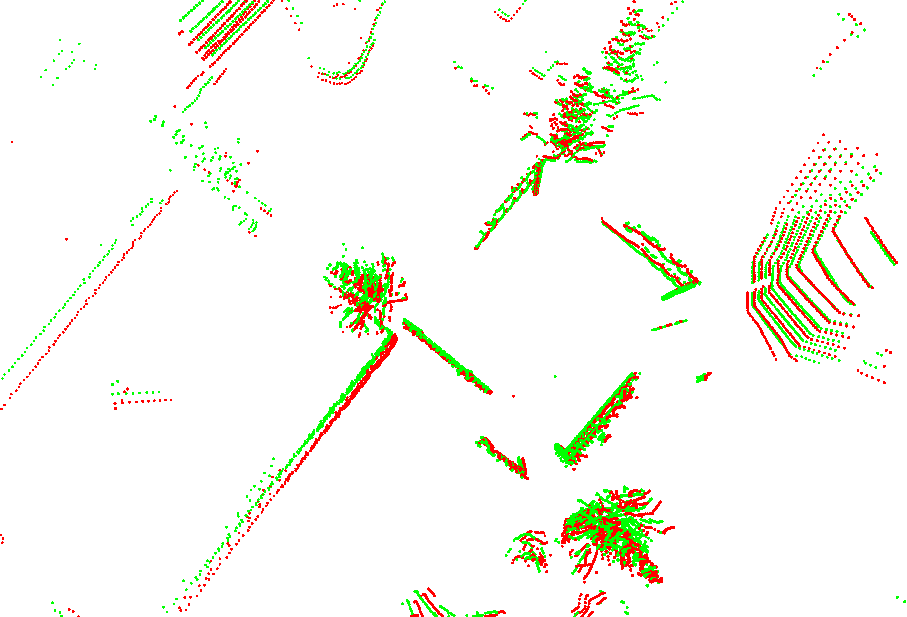} \\*[2pt]
    \rotatebox{90}{\hspace{2.8em}MAC} & 
    \imshowmac{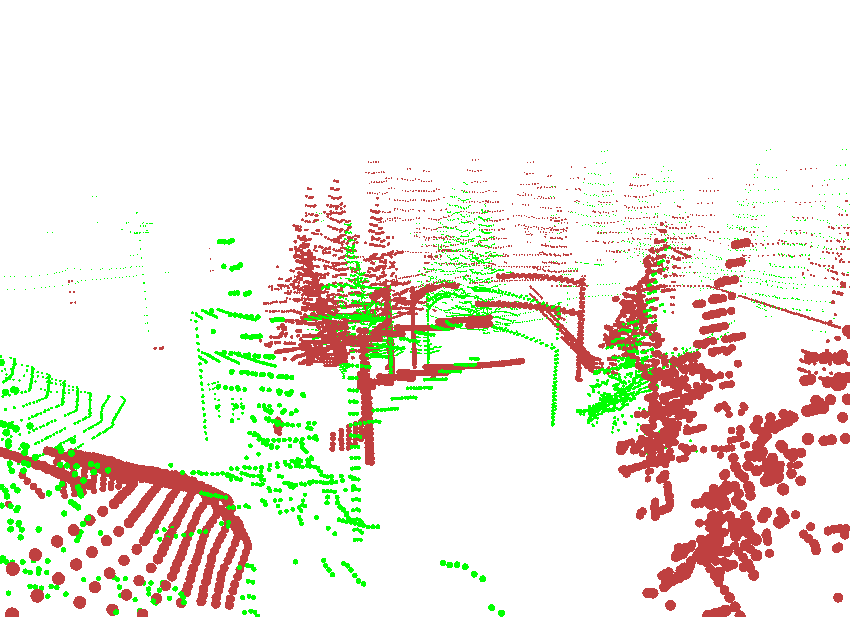} & 
    \imshowmac{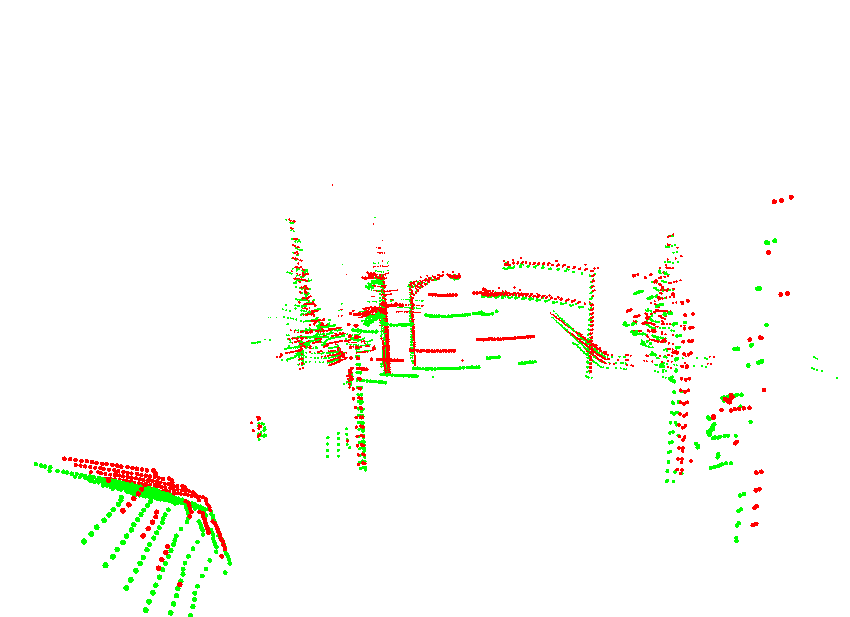} &  
    \imshowmac{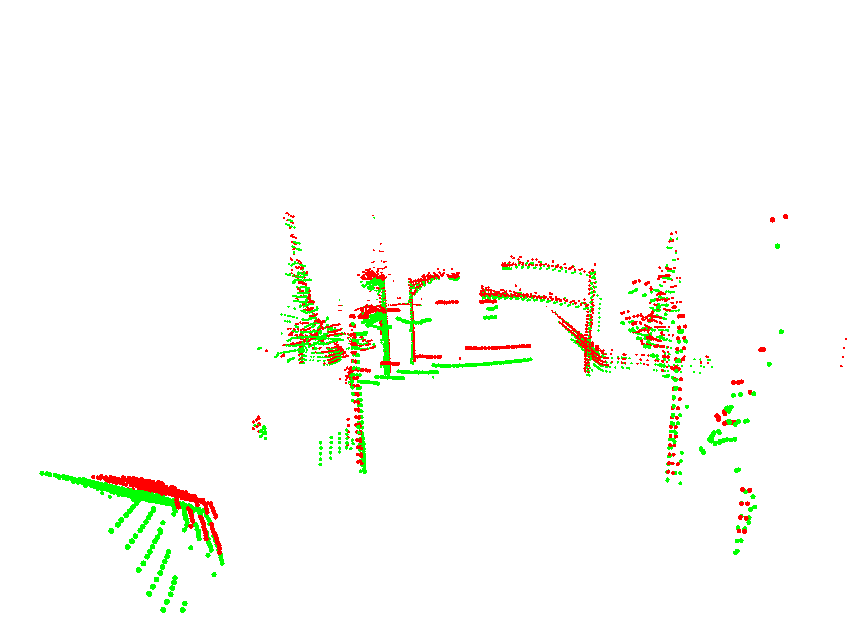} &
    \imshowmac{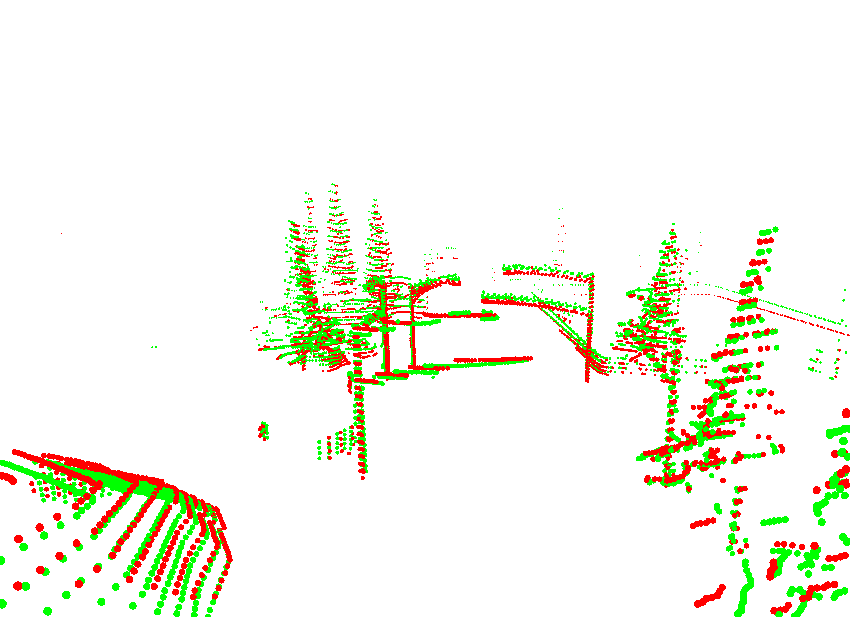} \\
    \end{tabular}
    \vspace{0.7em}
    \caption{Qualitative results of point cloud registration on point clouds of example test scenes, under different optimization schemes. From left to right: optimizing the LiDAR only, optimizing the registration hyperparameters only, performing joint optimization.  Only the relevant parts are shown for visualization purposes.}
    \label{fig:qualitative_results}
\end{figure*}

\subsection{Parameter analysis}

\begin{figure*}[!t]
\footnotesize
\centering
\includegraphics[width=1.0\linewidth]{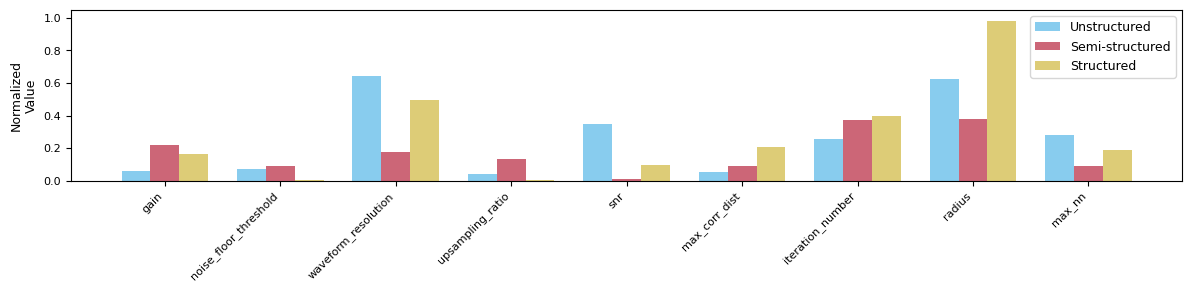}
\vspace{0.1em}
\caption{Jointly optimized LiDAR and FGR parameters vary across different scene types: unstructured, semi-structured, and structured.}
\label{fig:param_analysis}
\end{figure*}

\Cref{fig:param_analysis} shows how the optimal configuration of LiDAR sensing and registration parameters depends on the type of environment.
The optimized parameters reveal consistent trends across scene types, reflecting adaptations in both LiDAR sensing and FGR registration. The \textit{gain} is lowest in unstructured scenes and increases with structure, indicating the need for greater signal amplification in predictable environments. Conversely, the \textit{noise floor threshold} increases from structured to semi-structured scenes, suggesting adaptive filtering in response to ambient variability.

\textit{Waveform resolution} is highest in unstructured scenes, consistent with the importance to resolve fine geometric details, while structured scenes operate with reduced resolution and lower \textit{upsampling ratio} and \textit{SNR}, favoring signal stability over detail.

In the registration pipeline, \textit{maximum correspondence distance} is optimized to larger values for structured scenes, reflecting how regular geometry allows for more lenient matching. In contrast, unstructured scenes require stricter correspondence thresholds. \textit{FGR iteration count} is larger in structured scenes, likely due to the need for finer convergence in repetitive or symmetric geometries. Neighborhood parameters used in FPFH, such as \textit{search radius} and \textit{max nearest neighbors}, are set to high values in unstructured scenes, emphasizing the need for richer local geometric context. Overall, these findings align with theoretical expectations: less structured scenes benefit from higher-fidelity sensing and tighter registration constraints, while structured scenes exploit geometric regularity to permit broader matching and deeper iterative refinement.

\section{Discussion}

The main result of this paper is a demonstration that jointly optimizing the parameters of the LiDAR and registration algorithm yields improved performance over optimizing them individually. This shows the importance of considering the entire imaging pipeline for 3D data processing, both confirming the findings in \cite{goudreault2023lidarhyperoptim} and extending them to the case of point cloud registration. Hence, adaptive pipelines could dynamically tune sensing and registration parameters to match the structural context, improving both robustness and generalizability in diverse environments.


A key challenge in deploying this method on real-world LiDAR hardware lies in the simulation-to-reality gap. While the simulator developed by \cite{goudreault2023lidarhyperoptim} incorporates full wavefront processing alongside a realistic DSP model, it still falls short in capturing critical aspects such as stochastic sensor noise, occlusions, and weather-induced distortions. Furthermore, the CARLA engine assumes perfect sensor calibration and ego-pose estimation, whereas real-world acquisitions are subject to GPS/IMU drift, rolling shutter artifacts, and hardware-specific imperfections. These discrepancies limit the direct transferability of optimized LiDAR parameters from simulation to hardware. We hope that this work encourages LiDAR manufacturers to provide greater transparency into their processing pipelines and fosters the development of more realistic simulators, thereby enabling deeper exploration of the role of the complete 3D imaging and data pipeline in downstream computer vision tasks.

{\small
\paragraph*{Acknowledgments}

This research was supported by NSERC grant ALLRP 580274-22 and Bentley Systems. We thank Harshitha Devendra for assistance with figure preparation and Yannick Hold-Geoffroy for providing insightful comments that helped improve the manuscript.}

\label{sec:discussion}


\bibliography{references}
\end{document}